\documentclass[runningheads]{llncs}
\usepackage{graphicx}
\usepackage{amsmath,amssymb} %
\usepackage{color}
\usepackage{pgfplots}
\usepackage{tikz}
\usepackage{array}
\usepackage{multirow}
\usepackage{booktabs}
\usepackage{colortbl}
\usepackage{floatrow}
\usepackage[hyphens]{url}
\usepackage{xcolor}
\usepackage[bookmarks=false]{hyperref}
\hypersetup{
    colorlinks,
    linkcolor={blue!50!black},
    citecolor={blue!50!black},
    urlcolor={blue!80!black}
}

\DeclareMathOperator*{\argmax}{argmax}
\DeclareMathOperator*{\argmin}{argmin}
\newcommand\myparagraph[1]{\vspace{8pt}\noindent\textbf{#1}\quad}
\newcommand\etal{\textit{et al.}}

\newcommand\sectionname{Sect.}

\begin{document}
\pagestyle{headings}
\mainmatter

\title{Part Detector Discovery in Deep Convolutional Neural Networks }
     
\titlerunning{Part Detector Discovery in Deep Convolutional Neural Networks}
\authorrunning{Marcel Simon \and Erik Rodner \and Joachim Denzler}

\author{Marcel Simon \and Erik Rodner \and Joachim Denzler}
\institute{Computer Vision Group, Friedrich Schiller University of Jena, Germany\\
\url{www.inf-cv.uni-jena.de}}

\maketitle

\begin{abstract}
Current fine-grained classification approaches often rely
on a robust localization of object parts to extract 
localized feature representations suitable for discrimination.
However, part localization is a
challenging task due to the large variation of appearance and pose.
In this paper, we show how pre-trained convolutional neural networks 
can be used for robust and efficient object part discovery and localization without the
necessity to actually train the network on the current dataset. Our approach called
``part detector discovery'' (PDD)
is based on analyzing the gradient maps of the network outputs and finding
activation centers spatially related to annotated semantic parts or bounding boxes. 

This allows us not just to obtain excellent performance on the CUB200-2011 dataset,
but in contrast to previous approaches also to perform detection and bird classification jointly 
without requiring a given bounding box annotation during testing and ground-truth parts during training.
The code is available at \url{http://www.inf-cv.uni-jena.de/part_discovery} and \url{https://github.com/cvjena/PartDetectorDisovery}.
\end{abstract}

\section{Introduction}

In recent years, the concept of \textit{deep learning} has gained tremendous interest 
in the vision community. One of the key ideas is to jointly learn a model for the whole classification
pipeline from an input image to the final outputs. A successful model especially for classification are deep convolutional
neural networks (CNN) \cite{krizhevsky2012imagenet}. A CNN model can be seen as the concatenation of
several processing steps similar to algorithmic steps in previous ``non-deep'' recognition models.
The steps sequentially transform
the given input into a likely more abstract representation
\cite{bengio2013deep,bengio2009learning,zeiler2013visualizing} and finally
to the expected output. 

Due to the large number of free parameters, deep models usually need to be learned from large-scale data, such as
the ImageNet dataset used in \cite{krizhevsky2012imagenet}, and learning is a computationally demanding step. 
The very recent work of \cite{donahue2013decaf,sermanet14overfeat,razavian2014cnn} shows that pre-trained deep models
can also be exploited  for datasets which they have not been trained on. In particular, 
efficient and very powerful feature representations
can be obtained that lead to a significant performance improvement on several vision benchmarks. 
Our work follows a similar line of thought and in particular the question 
we were interested in is: \textit{``Can we re-use pre-trained deep
convolutional networks for part discovery and detection? Does a deep model 
learned on ImageNet already include implicit detectors related
to common parts found in fine-grained recognition tasks?''}

The answer to both questions is yes and to show this we present a novel 
part discovery and detection scheme using pre-trained deep convolution neural networks.
Object representations for visual recognition are often part-based~\cite{dpm,berg2013partclass} 
and in several cases information about semantic parts is given. The parts
of a bird, for example, include the belly, the wings, the beak and the tail. 
The benefit of part-based representations is especially notable in fine-grained classification tasks, in which the objects of different categories are similar in
shape and general appearance, but differ greatly in some small parts. This is also the
application scenario considered in our paper.

\begin{figure}[tbp]
\begin{center}
  \includegraphics[width=0.73 \textwidth]{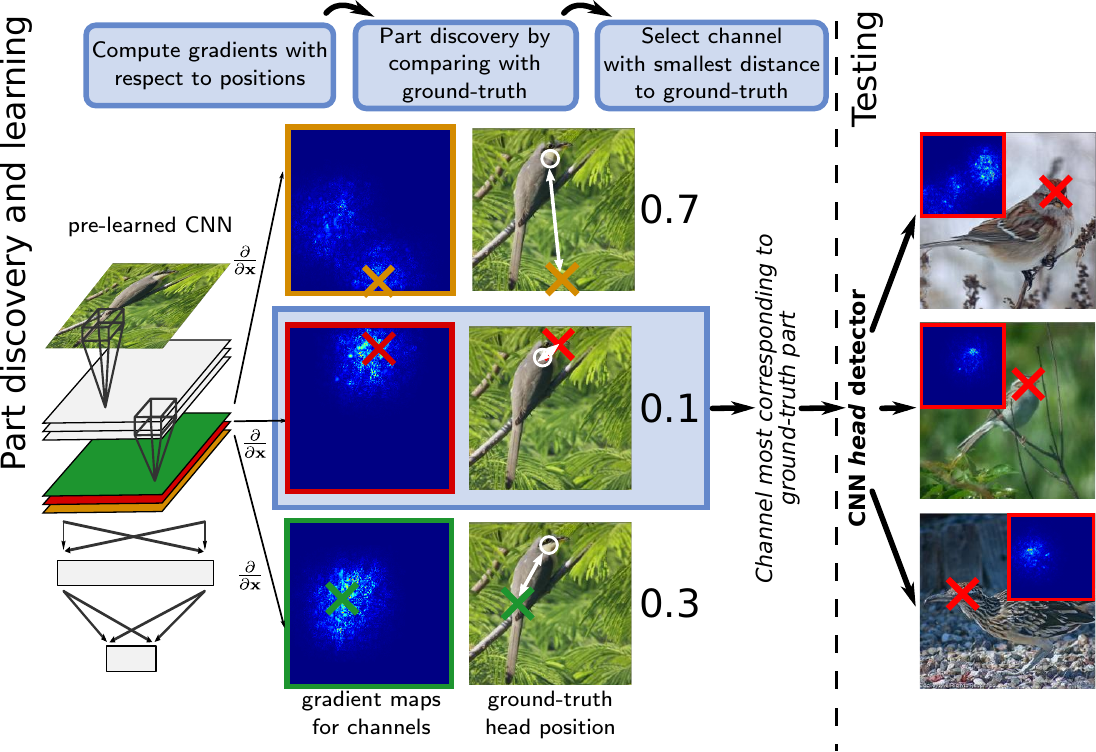}
\end{center}
\vspace{-0.3cm}
\caption{Outline of our approach during learning using the \textit{part strategy}: (1) compute gradients of CNN channels with respect to image positions, (2) estimate activation centers,
    (3) find spatially related semantic parts to select useful channels that act as part detectors later on.}\label{fig:teaser} 
\end{figure}

Our technique for providing such a part-based representation is based on computing gradient maps with respect to certain 
channel outputs and finding clusters of high activation within. 
This is followed by selecting channels which have their corresponding clusters closest to ground-truth positions of semantic parts or bounding boxes. An outline of our
approach is given in \figurename~\ref{fig:teaser}. The most interesting aspect is that after finding these associations, 
parts can be reliably detected without much additional computational effort from the results of the deep CNN.

\section{Related Work}

\myparagraph{CNN for Classification}
The approach of Krizhevsky~\etal~\cite{krizhevsky2012imagenet}
demonstrates the capabilities of multilayered architectures for image classification
by achieving an outstanding error rate on the LSVRC2012 dataset. 
Donahue~\etal~\cite{donahue2013decaf} use a CNN with a similar architecture and analyze how well a network
trained on ImageNet performs on other not related image classification datasets. 
They use the output of a hidden layer
as feature representation followed by a SVM \cite{cortes1995svm} learned on the new dataset. 
The authors have also published ``DeCAF'', an open source framework for CNNs~\cite{donahue2013decaf}, which we also use in our work. 
Similar studies have been done by \cite{sermanet14overfeat,razavian2014cnn}. In contrast to
their work, we focus on part discovery with pre-trained CNNs rather than classification.

An important aspect of \cite{zeiler2013visualizing}
is the qualitative analysis of the learned intermediate representations. In particular, they show
that CNNs trained on ImageNet contain models for basic object shapes at lower layers and object part models
at higher layers. This is exactly the property that we make use of in our approach.
Although the CNN we use was not particularly trained on the fine-grained task we consider, it can be
used for basic shape and part detection.

\myparagraph{CNN for Object Detection}
Motivated by the good results of CNNs in classification, several methods were proposed to
apply this technique to object detection. Erhan~\etal~\cite{erhan14cnnbboxpred} 
use a deep neural network to directly predict
the coordinates of a fixed number of bounding boxes and the
corresponding confidence. They determine the category of the object in 
each bounding box using a second CNN. A main drawback of this approach is
the fixed number of bounding box proposals once the network has been trained.
In order to change the number of proposals, the whole networks needs to be trained again.
An alternative to the direct prediction of the bounding box coordinates is presented in
\cite{szegedy13cnndetection}. They train a deep neural network to predict
an object mask at multiple scales. While the predicted mask is similar to our gradient maps,
they need to train an additional CNN specifically for the object detection task. 
Simonyan~\etal~\cite{simonyan2013deep} perform object localization by 
analyzing the gradient of a deep CNN trained for classification.
They compute the gradient of the winning class score with respect to
the input image. By thresholding the absolute gradient values,
seed background and seed foreground pixels are located and used as initialization
for the GrabCut segmentation algorithm. 
We make use of their idea to use gradients. However, while \cite{simonyan2013deep} use the 
gradients for segmentation, we use them for part discovery.
In addition to this, we introduce the idea of using intermediate layers and techniques like 
the aggregation of gradients of the same channel.

\myparagraph{CNN for Part Localization}
Many classification approaches in fine-grained recognition use part-based object models and 
hence also require part detection. At the moment, most part-based models use low-level 
features like histogram of oriented gradients \cite{hog} for detection and description. 
Examples are the deformable part model (DPM) \cite{dpm}, regionlets \cite{regionlets}, and
poselets \cite{poselets}. Facing the success of features from deep CNN, a current line of research
is the use of these features as a replacement for the low-level features of the part-based models.
For example, \cite{donahue2013decaf} relies on deformable part descriptors \cite{zhang2013dpd}, which is inspired by
DPM. While the localization is still done with low-level features,
the descriptor is calculated using the activations of a deep CNN for each predicted part.
Zhang~\etal~\cite{zhang14panda} use poselets for part discovery as well as detection and calculate features for each region using a deep CNN. 
While DPM and regionlets work well on many datasets, they face problems
on datasets with large intraclass variance and unconstrained settings especially due to 
the low-level features that are still used for localization. In contrast, our approach 
implicitly exploits all high-level features that a large CNN has learned already. It also allows us
to solve part localization and the consequent classification within the same framework. 

Entirely replacing the low-level features is difficult, because a CNN produces global
features. Zou~\etal~\cite{zou14cnnfeaturesregionlets} solves this by associating the output of 
a hidden layer with spatially related parts of the input. This allows 
them to apply the regionlets approach. However, 
their approach requires numerous evaluations of the CNN and the features are not arbitrarily dense.
In contrast, our approach is working on the full resolution of the input. 

The work of Jain~\etal~\cite{jain14cnnPose} uses the sliding window approach for part localization with CNNs. They evaluate a
CNN at each position of the window in order to detect human body parts in images without using bounding box annotations. 
This requires hundreds or thousands of CNN evaluations. 
As sufficiently large CNNs take some time for prediction,
this results in long run times. In contrast, only one forward- and one back-propagation
pass per part is required for the localization with our approach.
In \cite{toshev14deeppose} and \cite{sun13cnnfacial}, the part positions are estimated by a CNN
specifically trained for this task. The outputs of their network are the coordinates of each part.
In contrast to their work, our approach does not require any separately trained
neural network but can exploit a CNN already trained for a different large-scale classification task.

\section{Localization with Deep Convolutional Neural Networks}

In the following, we briefly review the concept of deep CNNs in Sect. \ref{sec:deep_arch} 
and the use of gradient maps for object localization
presented by Simonyan~\etal~\cite{simonyan2013deep} in Sect. \ref{sec:gradient_localization}
as these are the ideas our work is based on.

\subsection{Deep Convolutional Architectures \label{sec:deep_arch}}

\begin{figure}[tbp]
  \centering
  \includegraphics[width=0.9\textwidth]{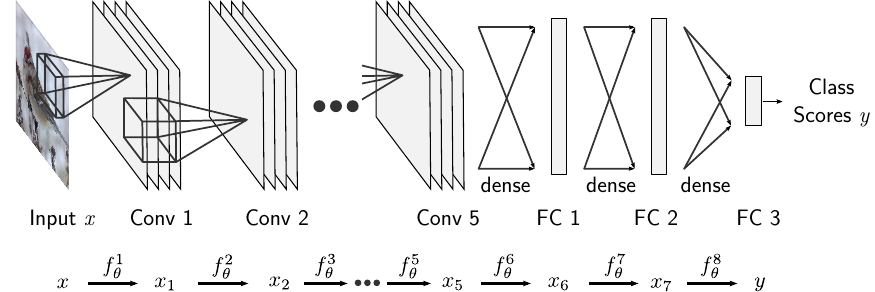}
\caption{Example of a convolutional neural network used for classification. Each convolutional layer
convolves the output of the previous layer with multiple learned filter masks, applies an element-wise non-linear activation function, and optionally
combines the outputs by a pooling operation. The last two layers
are fully connected layers and multiply the input with a matrix of learned parameters followed
by a non-linear activation function. The output of the network are scores for each of the learned categories. \label{fig:cnn_architecture} }
\end{figure}

A key idea of deep learning approaches is that the whole classification pipeline
consists of one combined and jointly trained model. Many non-deep systems
rely on hand-crafted features and separately trained machine learning algorithms.
In contrast, no hand-crafted features are required in deep learning 
architectures as they are automatically learned from the training data and 
no separately trained machine learning algorithm is required as the deep
architecture also covers this part. 
Most recent deep learning architectures for vision are based on a single CNN. 
CNNs are feed forward neural networks, which concatenate
several layers of different types. These layers are often related to techniques that a lot of non-deep approaches
for visual recognition used without joint parameter tuning:
\begin{enumerate}
\item Convolutional layers perform filtering with multiple filter masks, which is related to computing the distance
of local features to a given codebook in the common bag-of-words (BoW) pipeline.
\item Pooling layers spatially combine filter outputs, which is related to classical spatial pyramid matching~\cite{lazebnik2006beyond}.
\item Non-linear layers, such as the rectified linear unit used by \cite{krizhevsky2012imagenet}, are related to non-linear encoding techniques for BoW~\cite{coates2011importance}.
\item Several fully connected layers at the end act as nested linear classifiers. 
\end{enumerate}
The final output of the CNN are scores for each object category. The architecture
is visualized in a simplified manner in \figurename~\ref{fig:cnn_architecture}. For details about the network structure
we refer to \cite{krizhevsky2012imagenet} and \cite{donahue2013decaf} since we use the same pre-trained CNN.

As visualized in the figure, we will denote the transformation performed by the CNN
as a function $f_\theta(x)$, which maps the input image $x$ to the classification
scores for each category using the parameter vector $\theta$. We omit $\theta$ in the following
text since it is not relevant for our approach. The function $f$ is a concatenation
of functions $f_{}(x)=f_{}^{(n)}\circ f_{}^{(n-1)}\circ\dots\circ f_{}^{(1)}(x)$, where
the $f_{}^{(1)},f_{}^{(2)},\dots,f_{}^{(n)}$ correspond 
to the $n$ layers of the network. Furthermore, let $g_{}^{(k)}(x):=f_{}^{(k)}\circ f_{}^{(k-1)}\circ\dots\circ f_{}^{(1)}(x)$
be the output and $x^{(k)}:=g_{}^{(k-1)}(x)$ the input of the $k$-th layer.
Please note that $f^{(j)}$ represents only the transformation of layer $j$, while
$g^{(j)}$ includes all transformations from input image to layer $j$. 

Furthermore and most importantly for our approach, the output of the first layers is organized in channels with each
element in the channel being related to a certain area in the input image. The outputs of a channel are the result of several nested
convolutions, non-linear activations and pooling operations applied to the input image. Therefore, we can view them as results of pooled detector scores, 
a connection that we make use of for our part discovery scheme in \sectionname~\ref{sec:part_detection}.

\subsection{Localization Using Gradient Maps\label{sec:gradient_localization}}
Deep convolutional neural networks trained for classification can also be used
for other visual recognition tasks like object localization. Simonyan~\etal~\cite{simonyan2013deep} 
present a method, which calculates the gradient of the winning class score with respect to the 
input image. The intuition is that important foreground pixels
have a large influence on the classification score and hence have a large absolute gradient value.
Background pixels, however, do not influence the classification result and hence have a small gradient.
The experiments in \cite{simonyan2013deep} support this intuition and it allows them to use this information for segmentation of
foreground and background. The next paragraph briefly reviews 
the calculation of the gradients as we also use gradient maps in our system.

The  calculation of the gradient $\frac{\partial f_i}{\partial x}$ of the model output $(f(x))_i$ 
with respect to the input image $x$ is  
similar to the back-propagation of the classification 
error during training. Using the notation introduced in Sect. \ref{sec:deep_arch}, 
the gradient can be computed using the chain rule as
\begin{align}
\label{eq:jacobian1}
\frac{\partial f_{i}}{\partial x} &= \frac{\partial f_{i}}{\partial g_{}^{(n-1)}}\frac{\partial g_{}^{(n-1)}}{\partial g_{}^{(n-2)}}\dots\frac{\partial g_{}^{(2)}}{\partial g_{}^{(1)}}\frac{\partial g_{}^{(1)}}{\partial x} = \frac{\partial g_{i}^{(n)}}{\partial x^{(n)}}\frac{\partial g_{}^{(n-1)}}{\partial x^{(n-1)}}\dots\frac{\partial g_{}^{(1)}}{\partial x}.
\end{align}
Each factor $\frac{\partial g_{}^{(j)}}{\partial x^{(j)}}$ of this product is a Jacobian matrix 
of the output $g^{(j)}$ with respect to the input $x^{(j)}$ of layer $j$ . This means that
the gradient can be computed layer by layer. These factors are also calculated during back-propagation. 
The only difference is that the last derivative is with respect to 
$x$ instead of $\theta$.

\begin{figure}[tbp]
\begin{centering}
\resizebox{\textwidth}{!}{%
%
%
{
\pgfplotsset{ticks=none}
\begin{tikzpicture}

\begin{axis}[%
width=3.5cm,
height=3.5cm,
axis on top,
scale only axis,
xmin=0.5,
xmax=227.5,
y dir=reverse,
ymin=0.5,
ymax=227.5,
name=plot2
]
\addplot [forget plot] graphics [xmin=0.5,xmax=227.5,ymin=0.5,ymax=227.5] {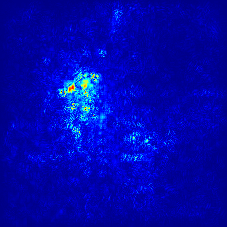};
\end{axis}

\begin{axis}[%
width=3.5cm,
height=3.5cm,
axis on top,
scale only axis,
xmin=0.5,
xmax=227.5,
y dir=reverse,
ymin=0.5,
ymax=227.5,
at=(plot2.left of south west),
anchor=right of south east
]
\addplot [forget plot] graphics [xmin=0.5,xmax=227.5,ymin=0.5,ymax=227.5] {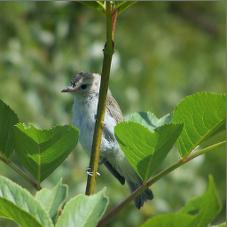};
\end{axis}

\begin{axis}[%
width=3.5cm,
height=3.5cm,
axis on top,
scale only axis,
xmin=0.5,
xmax=227.5,
y dir=reverse,
ymin=0.5,
ymax=227.5,
name=plot3,
at=(plot2.right of south east),
anchor=left of south west
]
\addplot [forget plot] graphics [xmin=0.5,xmax=227.5,ymin=0.5,ymax=227.5] {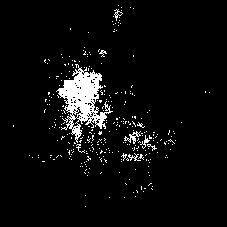};
\end{axis}

\end{tikzpicture}%
}
  \par\end{centering}
  \begin{centering}
  \smallskip{}
%
%
\begin{tikzpicture}

\begin{axis}[%
width=3.5cm,
height=3.5cm,
axis on top,
scale only axis,
xmin=0.5,
xmax=227.5,
y dir=reverse,
ymin=0.5,
ymax=227.5,
hide axis,
name=plot2
]
\addplot [forget plot] graphics [xmin=0.5,xmax=227.5,ymin=0.5,ymax=227.5] {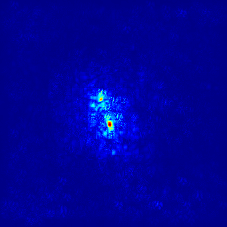};
\end{axis}

\begin{axis}[%
width=3.5cm,
height=3.5cm,
axis on top,
scale only axis,
xmin=0.5,
xmax=227.5,
y dir=reverse,
ymin=0.5,
ymax=227.5,
hide axis,
at=(plot2.left of south west),
anchor=right of south east
]
\addplot [forget plot] graphics [xmin=0.5,xmax=227.5,ymin=0.5,ymax=227.5] {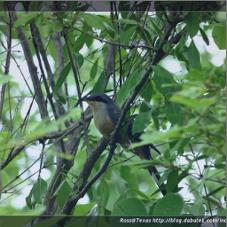};
\end{axis}

\begin{axis}[%
width=3.5cm,
height=3.5cm,
axis on top,
scale only axis,
xmin=0.5,
xmax=227.5,
y dir=reverse,
ymin=0.5,
ymax=227.5,
hide axis,
at=(plot2.right of south east),
anchor=left of south west
]
\addplot [forget plot] graphics [xmin=0.5,xmax=227.5,ymin=0.5,ymax=227.5] {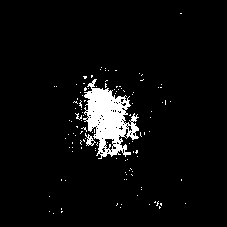};
\end{axis}
\end{tikzpicture}%
}
\par\end{centering}
\caption{Examples for the gradient gradient maps (continuous and thresholded with the $95\%$-quantile) that are calculated with respect to the winning class score.  \label{fig:Examples_Gradient} }
\end{figure}

The gradient $ \frac{\partial f_i}{\partial x}$  can be reshaped
such that it has the same shape as the input image. 
The result is called gradient map and \figurename~\ref{fig:Examples_Gradient}
visualizes the gradient maps that are calculated with respect to the winning class score for two examples. At first glance,
a gradient might seem very similar to a saliency map. However, while a saliency map captures objects that distinguish
themselves from the background \cite{borji13saliency}, the gradient values are only
large for pixels, that influence the classification result. 
Hence, conspicuous background objects will be highlighted in a salient map but not in a gradient map as they do not influence the 
score of the winning class. The second image
in \figurename~\ref{fig:Examples_Gradient} is a good example, whose strong
background patterns are not highlighted in the gradient map.

\section{Part Discovery in CNNs by Correspondence \label{sec:part_detection}}
Recent fine-grained categorization experiments have shown that the
location of object parts is a valuable information and allows to boost
the classification accuracy significantly \cite{goering2014parttransfer,berg2013partclass,donahue2013decaf}. However, 
the precise and fast localization of parts can
be a challenging task due to the great variety of poses in some datasets.

In Sect. \ref{sec:part_localization}, we present a novel approach for automatically detecting parts of
an object. Given a test image of an object from a previously known set
of categories, the algorithm will detect visible parts and locate
them in the image. This is followed by Sect. \ref{sec:part_classification} presenting a part-based classification approach,
which is used as an application of our part localization method in the experiments.

\subsection{Part Localization\label{sec:part_localization}}

\myparagraph{Gradient Maps for Part Discovery}
Like the object localization system of the previous section, our algorithm 
requires a pre-trained CNN trained for image 
classification. The classification task it was trained for does not need to be 
directly related to the actual part localization task. For example, 
we used for our experiments the model of \cite{donahue2013decaf}, which
was trained on the ImageNet dataset. However, all our experiments are performed on
the Caltech Birds dataset. 

The part localization is done by calculating the gradient of the output elements
of a hidden layer with respect to the input image. Suppose the selected layer has $m$ output 
elements, then $m$ gradients with respect to the input image and hence $m$ gradient maps are 
computed. 
The calculation of the gradient maps for each element of a hidden layer is done in 
a similar way as the gradient of a class score. Let $b$ denote the chosen hidden layer.
Then, the gradient $\frac{\partial g^{(b)}_j}{\partial x}(x)$ of the $j$-th element of
 layer $b$ with respect to the input image $x$ is calculated as
\begin{eqnarray}
\frac{\partial g^{(b)}_j}{\partial x} & = & \frac{\partial g_{j}^{(b)}}{\partial g_{}^{(b-1)}}\frac{\partial g_{}^{(b-1)}}{\partial g_{}^{(b-2)}}\dots\frac{\partial g_{}^{(1)}}{\partial x} = \frac{\partial g_{j}^{(b)}}{\partial x^{(b)}}\cdot\frac{\partial g_{}^{(b-1)}}{\partial x^{(b-1)}}\dots\frac{\partial g_{}^{(1)}}{\partial x}\,.
\end{eqnarray}
As before, we can also make use of the back-propagation scheme for gradients
already implemented in most CNN toolboxes.
The gradient maps of the same channel are added pixelwise in order to
obtain one gradient map per channel. The intuition is that 
each element of a hidden layer is sensitive to a specific pattern in the input image.
All  elements of the same channel are sensitive to the same pattern 
but are focused on a different location of this pattern.

\myparagraph{Part Discovery by Correspondence}
We now want to identify channels, which are related to object parts. 
In the following, we assume that the ground-truth part locations $z_i$ of the training images $x_i$ 
are given. However, our method can be also provided with the location of the bounding box only as 
we will show later. We associate a binary latent variable $h_k$ with
each channel $k$, which indicates whether the channel is related to an object part.
Our part discovery scheme can be motivated as a maximum likelihood estimation of these variables. 
First, let us consider the task of selecting the most related channel corresponding to a part which
can be written as (assuming $x_i$ are independent samples):
\begin{align}
\hat{k}&=\operatorname{argmax}\limits_{1 \leq k \leq K} \;p(\vec{X}\;|\;h_k=1) = \argmax_{1 \leq k \leq K} \prod_{i=1}^{N} \frac{p\left(h_k=1|x_{i}\right) p\left(x_{i}\right)}{p\left(h_k=1\right) }\,.
\end{align}
where $\vec{X}$ is the training data and $K$ is the total number of channels. In the following, we assume a flat prior for $p(h_k=1)$ and $p(x_i)$. The term $p\left(h_k=1|x_{i}\right)$ expresses the probability that channel $k$ corresponds to the part currently under consideration given a single
training example $x_i$. This is the case when the position $p_i^k$ estimated using channel $k$ equals the ground-truth part position $z_i$. However, the estimated position $p_i^k$ is likely not perfect and we assume it to be a Gaussian random variable distributed as $p_i^k \sim \mathcal{N}(\mu^k_i, \sigma^2)$, where $\mu^k_i$ is a center of activation extracted from the gradient map of channel $k$. We therefore have:
\begin{align}
p(h_k=1 | x_i) &= p( p_i^k = z_i | x_i) = \mathcal{N}( z_i | \mu^k_i, \sigma^2 )
\end{align}
Putting it all together, we obtain a very simple scheme for selecting a channel:
\begin{align}
\hat{k}&=\argmax_{1 \leq k \leq K} \sum\limits_{i=1}^N \log p(h_k=1 | x_i)
       =\argmin_{1 \leq k \leq K} \sum\limits_{i=1}^N \| \mu^k_i - z_i \|^2
\end{align}
For all gradient maps of all training images, the center of activation $\mu_i^k$
is calculated as explained in the subsequent paragraph. These locations
are compared to the ground-truth part locations $z_i$ by computing the mean
distance. Finally, for each ground-truth part, the channel with the
smallest mean distance is selected. The result is a set of channels,
which are sensitive to different parts of the object. There does not 
need to be a one-to-one relationship between parts and channels,
because the neural network is not trained on semantic parts. We refer to this method as \textit{part strategy}. 

\myparagraph{Part Discovery without Ground-truth Parts}
In many scenarios, ground-truth part annotations are not available. Our approach 
can also be applied in these cases by selecting relevant channels based on the 
bounding box of the training images only. 
We evaluate two different approaches related to different models for $p(h_k=1|x_i)$.
First, we count for every channel 
how often the activation center is within the bounding box $\text{BB}(x_i)$ of the image $x_i$. We select 
the channels with the highest count, as these are most likely to correspond to the object of interest.
It can be shown that this \emph{counting strategy} is related to the following model for arbitrary $0<\epsilon<1$:
\begin{align}
p(h_k=1|x_i) &= \begin{cases} 1-\epsilon & \mu_i^k \in \text{BB}(x_i)\\ \epsilon & \text{otherwise} \end{cases}
\end{align}
Second, we extend the first approach by taking the distance to the \emph{bounding box} border into account.
If the activation center of a channel is within the bounding box of a training image, the cost is 0.
If it is outside, the cost equals to the Euclidean distance to the bounding box border.

\myparagraph{Finding Activation Centers}
The assumption for finding the center of activation $\mu_i$ is that high absolute gradient values
are concentrated in local areas of the gradient maps and these areas correspond
to certain patterns in the image. 
In order to robustly localize the center point
of the largest area, we fit a Gaussian mixture model with $K$ components to the
pixel locations weighted by the normalized absolute gradient values. 
We then take the mean location of the most prominent cluster in the mixture
as the center of activation. In comparison to simply taking the maximum position
in the gradient map, this approach is much more robust to noise as we show in the experiments. 

\myparagraph{Part Detection}
In order to locate parts in a new image, the gradient maps for 
all selected channels are calculated. The activation center
is estimated and returned as the location
of the part. If the gradient map is equal to $\mathbf{0}$, the part is likely not visible
in the image and marked as occluded.
\begin{figure}[tbp]
\resizebox{0.49\textwidth}{!}{%
\begin{centering}
%
%
{
\pgfplotsset{ticks=none}
\begin{tikzpicture}

    \tikzstyle{every node}=[font=\LARGE]
\begin{axis}[%
width=3.5cm,
height=3.5cm,
axis on top,
scale only axis,
xmin=0.5,
xmax=227.5,
xticklabel pos=right,
xlabel={Body},
xlabel near ticks,
y dir=reverse,
ymin=0.5,
ymax=227.5,
name=plot2
]
\addplot [forget plot] graphics [xmin=0.5,xmax=227.5,ymin=0.5,ymax=227.5] {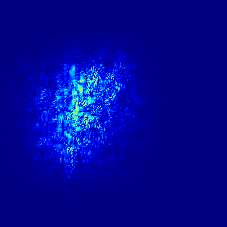};
\end{axis}

\begin{axis}[%
width=3.5cm,
height=3.5cm,
axis on top,
scale only axis,
xmin=0.5,
xmax=227.5,
xticklabel pos=right,
xlabel={Input image},
xlabel near ticks,
y dir=reverse,
ymin=0.5,
ymax=227.5,
at=(plot2.left of south west),
anchor=right of south east
]
\addplot [forget plot] graphics [xmin=0.5,xmax=227.5,ymin=0.5,ymax=227.5] {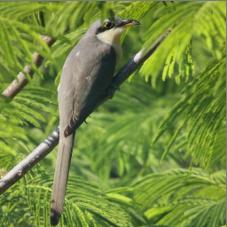};
\end{axis}

\begin{axis}[%
width=3.5cm,
height=3.5cm,
axis on top,
scale only axis,
xmin=0.5,
xmax=227.5,
xticklabel pos=right,
xlabel={Head},
xlabel near ticks,
y dir=reverse,
ymin=0.5,
ymax=227.5,
name=plot3,
at=(plot2.right of south east),
anchor=left of south west,
inner sep=0.7em]
\addplot [forget plot] graphics [xmin=0.5,xmax=227.5,ymin=0.5,ymax=227.5] {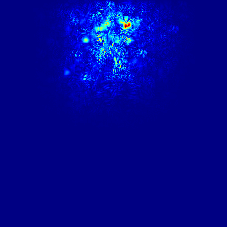};
\end{axis}

\begin{axis}[%
width=3.5cm,
height=3.5cm,
axis on top,
scale only axis,
xmin=0.5,
xmax=227.5,
xticklabel pos=right,
xlabel={Wings},
xlabel near ticks,
y dir=reverse,
ymin=0.5,
ymax=227.5,
at=(plot3.right of south east),
anchor=left of south west
]
\addplot [forget plot] graphics [xmin=0.5,xmax=227.5,ymin=0.5,ymax=227.5] {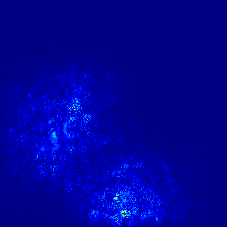};
\end{axis}
\end{tikzpicture}%
}
\end{centering}
}
\resizebox{0.49\textwidth}{!}{%
\begin{centering}
%
%
{ 
\pgfplotsset{ticks=none}
\begin{tikzpicture}

\tikzstyle{every node}=[font=\LARGE]
\begin{axis}[%
width=3.5cm,
height=3.5cm,
axis on top,
scale only axis,
xmin=0.5,
xmax=227.5,
xticklabel pos=right,
xlabel={Body},
xlabel near ticks,
y dir=reverse,
ymin=0.5,
ymax=227.5,
name=plot2
]
\addplot [forget plot] graphics [xmin=0.5,xmax=227.5,ymin=0.5,ymax=227.5] {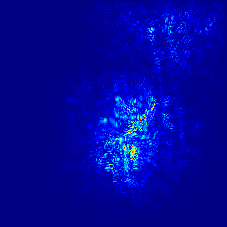};
\end{axis}

\begin{axis}[%
width=3.5cm,
height=3.5cm,
axis on top,
scale only axis,
xmin=0.5,
xmax=227.5,
xticklabel pos=right,
xlabel={Input image},
xlabel near ticks,
y dir=reverse,
ymin=0.5,
ymax=227.5,
at=(plot2.left of south west),
anchor=right of south east
]
\addplot [forget plot] graphics [xmin=0.5,xmax=227.5,ymin=0.5,ymax=227.5] {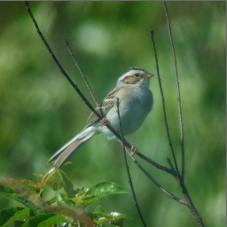};
\end{axis}

\begin{axis}[%
width=3.5cm,
height=3.5cm,
axis on top,
scale only axis,
xmin=0.5,
xmax=227.5,
xticklabel pos=right,
xlabel={Head},
xlabel near ticks,
y dir=reverse,
ymin=0.5,
ymax=227.5,
name=plot3,
at=(plot2.right of south east),
anchor=left of south west,
inner sep=0.7em
]
\addplot [forget plot] graphics [xmin=0.5,xmax=227.5,ymin=0.5,ymax=227.5] {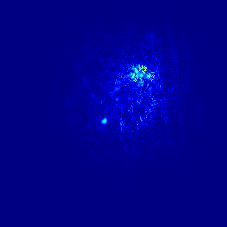};
\end{axis}

\begin{axis}[%
width=3.5cm,
height=3.5cm,
axis on top,
scale only axis,
xmin=0.5,
xmax=227.5,
xticklabel pos=right,
xlabel={Wings},
xlabel near ticks,
y dir=reverse,
ymin=0.5,
ymax=227.5,
at=(plot3.right of south east),
anchor=left of south west
]
\addplot [forget plot] graphics [xmin=0.5,xmax=227.5,ymin=0.5,ymax=227.5] {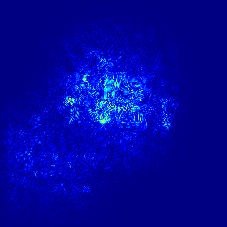};
\end{axis}
\end{tikzpicture}%
}
\end{centering}
}

\resizebox{0.49\textwidth}{!}{%
\begin{centering}
%
%
{
\pgfplotsset{ticks=none}
\begin{tikzpicture}

\begin{axis}[%
width=3.5cm,
height=3.5cm,
axis on top,
scale only axis,
xmin=0.5,
xmax=227.5,
y dir=reverse,
ymin=0.5,
ymax=227.5,
name=plot2
]
\addplot [forget plot] graphics [xmin=0.5,xmax=227.5,ymin=0.5,ymax=227.5] {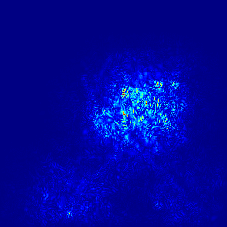};
\end{axis}

\begin{axis}[%
width=3.5cm,
height=3.5cm,
axis on top,
scale only axis,
xmin=0.5,
xmax=227.5,
y dir=reverse,
ymin=0.5,
ymax=227.5,
at=(plot2.left of south west),
anchor=right of south east
]
\addplot [forget plot] graphics [xmin=0.5,xmax=227.5,ymin=0.5,ymax=227.5] {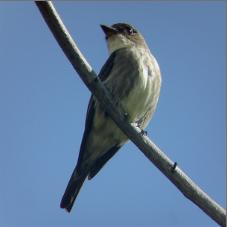};
\end{axis}

\begin{axis}[%
width=3.5cm,
height=3.5cm,
axis on top,
scale only axis,
xmin=0.5,
xmax=227.5,
y dir=reverse,
ymin=0.5,
ymax=227.5,
name=plot3,
at=(plot2.right of south east),
anchor=left of south west 
]
\addplot [forget plot] graphics [xmin=0.5,xmax=227.5,ymin=0.5,ymax=227.5] {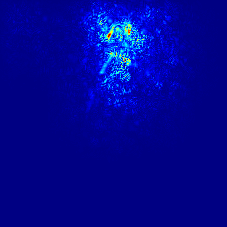};
\end{axis}

\begin{axis}[%
width=3.5cm,
height=3.5cm,
axis on top,
scale only axis,
xmin=0.5,
xmax=227.5,
y dir=reverse,
ymin=0.5,
ymax=227.5,
at=(plot3.right of south east),
anchor=left of south west
]
\addplot [forget plot] graphics [xmin=0.5,xmax=227.5,ymin=0.5,ymax=227.5] {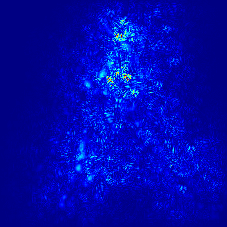};
\end{axis}
\end{tikzpicture}%
}
\end{centering}
}
\resizebox{0.49\textwidth}{!}{%
\begin{centering}
%
%
{ 
\pgfplotsset{ticks=none}
\begin{tikzpicture}

\begin{axis}[%
width=3.5cm,
height=3.5cm,
axis on top,
scale only axis,
xmin=0.5,
xmax=227.5,
y dir=reverse,
ymin=0.5,
ymax=227.5,
name=plot2
]
\addplot [forget plot] graphics [xmin=0.5,xmax=227.5,ymin=0.5,ymax=227.5] {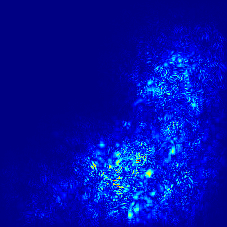};
\end{axis}

\begin{axis}[%
width=3.5cm,
height=3.5cm,
axis on top,
scale only axis,
xmin=0.5,
xmax=227.5,
y dir=reverse,
ymin=0.5,
ymax=227.5,
at=(plot2.left of south west),
anchor=right of south east
]
\addplot [forget plot] graphics [xmin=0.5,xmax=227.5,ymin=0.5,ymax=227.5] {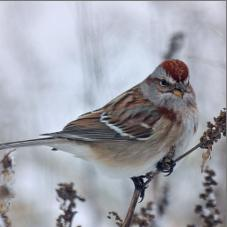};
\end{axis}

\begin{axis}[%
width=3.5cm,
height=3.5cm,
axis on top,
scale only axis,
xmin=0.5,
xmax=227.5,
y dir=reverse,
ymin=0.5,
ymax=227.5,
name=plot3,
at=(plot2.right of south east),
anchor=left of south west
]
\addplot [forget plot] graphics [xmin=0.5,xmax=227.5,ymin=0.5,ymax=227.5] {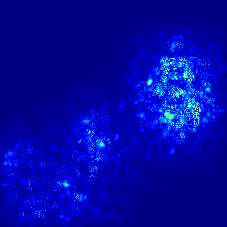};
\end{axis}

\begin{axis}[%
width=3.5cm,
height=3.5cm,
axis on top,
scale only axis,
xmin=0.5,
xmax=227.5,
y dir=reverse,
ymin=0.5,
ymax=227.5,
at=(plot3.right of south east),
anchor=left of south west
]
\addplot [forget plot] graphics [xmin=0.5,xmax=227.5,ymin=0.5,ymax=227.5] {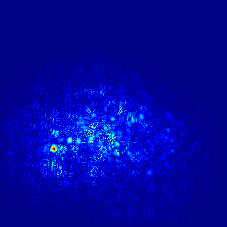};
\end{axis}
\end{tikzpicture}%
}
\end{centering}
}
\caption{Examples for gradient maps. Each group
of images  first shows the input image and then the gradient
maps of the channels associated with the body, head and the wings. A light blue
to red color means, that the gradient values of the corresponding
pixel is high. A deep blue corresponds to a gradient value of zero.
\label{fig:part_examples} }
\end{figure}
Figure \ref{fig:part_examples} visualizes the gradient maps for some test images
used in our experiments. For each image, three channels are shown which are
associated with semantic parts of a bird. In each group of images, the input
 is on the left followed by the normalized gradient maps of the channels associated with 
 the body, head and the wings. A light blue to red color corresponds to a large
absolute gradient value while a deep blue represents a zero gradient.

\myparagraph{Why should this work?}
The results of \cite{zeiler2013visualizing} suggest that at least
in the special case of deep CNNs, each element
of a hidden layer is sensitive to specific patterns in the image.
``Sensitive'' means in this case that the occurrence of a pattern
leads to a high change of the output. There is an implicit
association between certain image patterns and output elements of
a layer. 
In higher layers these patterns are increasingly abstract and hence
might correspond to a specific part of an object. 
Our method automatically identifies channels with this property.

As an example, suppose the input of a deep architecture is a RGB
color image and the first and only layer is a convolutional layer with two
channels representing different local edge filters. If the first filter
mask is a horizontal and the second one a vertical edge filter mask,
then the first channel is sensitive to horizontal edge patterns and
the second one to vertical edge patterns. This example also illustrates
that at least in the case of convolutional layers any output element
of the same channel reacts to the same pattern and that each element
just focuses on a different area in the input image. 
Of course, the example only discusses low-level patterns without any
direct connection to a specific part. However, experiments \cite{zeiler2013visualizing} indicate 
that in higher layers the patterns are more complex and correspond, for example,
to wheels or an eye.

\myparagraph{Implementation Details}
At first, it might seem that simply adding the gradients of all elements 
of a channel causes a loss of information. Negative and
positive gradients can cancel and might cause a close to zero gradient
for a discriminative pixel. Possibly adding the absolute gradient 
values seems to be  the better approach, but this is not the case.
First, since each element of a channel focuses  on a separate area in the
input image, the cancellation of positive and negative gradients is 
negligible. A small set of experiments supports this. Second,
calculating the sum of gradients requires only one back-propagation
call per channel instead of one call for each element of the channel.
In our experiments,  each channel has $36$ elements. Hence, 
there is a speedup of $36\times$ in our case.

The reason for this is the back-propagation algorithm which
directly calculates the sum of gradients if initialized correctly. 
We have already showed how to calculate the gradient $\frac{\partial g^{(b)}_j}{\partial x}$
of a element with respect to the input image. It is possible
to rewrite the calculation such that it shows more clearly how to 
apply the back-propagation algorithm. Let $e_j$ be the $j$-th unit vector.
Then 
\begin{eqnarray}
\frac{\partial g^{(b)}_j}{\partial x}= e_j \cdot \frac{\partial g^{(b)}}{\partial x} & = & e_j\cdot\frac{\partial g_{\theta}^{(b)}}{\partial g_{\theta}^{(b-1)}}\cdot\frac{\partial g_{\theta}^{(b-1)}}{\partial g_{\theta}^{(b-2)}}\cdot\dots\cdot\frac{\partial g_{\theta}^{(2)}}{\partial g_{\theta}^{(1)}}\cdot\frac{\partial g_{\theta}^{(1)}}{\partial x}(x)\label{eq:jacobian_equation}
\end{eqnarray}
where $\frac{\partial g^{(b)}}{\partial x}$ is the Jacobian matrix containing
the derivatives of each output element (the rows) with respect to each
input component of $x$ (the columns). Here, $e_j$ is the initialization for the
back-propagation at layer $b$ and all the other factors are applied 
during the backward pass. In other words, the initialization $e_j$ selects
the $j$-th row of the Jacobian matrix $\frac{\partial g^{(b)}}{\partial x}$.
Suppose there would be more than one element of $e_j$ equal to $1$. Then
the result is the sum of the corresponding gradient vectors. 
Let $s_{c}=\left(0,\dots,0,1,1,\dots,1,0,0,\dots,0\right) $
be such a ``modified'' $e_j$, where $c$ is the channel index. 
$s_c$ is $1$ at each position that corresponds to an element of channel $c$ and 
$0$ for all remaining components.
Replacing $e_j$ in Eq.~\ref{eq:jacobian_equation} by $s_c$ consequently 
calculates the required sum of gradients which belong to the same channel.
In contrast, the calculation of the sum of absolute gradients
values requires multiple backward passes with a new $e_j$ for each 
run. 

\subsection{Part-Based Image Classification \label{sec:part_classification}}
Part detection is only an intermediate step for most real world systems. We use the presented 
part discovery and detection method for part-based image classification. Our method adapts the
approach of  \cite{goering2014parttransfer} replacing the SIFT and color name features by 
the activations of a deep neural network and the non-parametric part transfer by the presented
part detection approach.

The feature vector consists 
of two types, the global and the part features. For the global feature extraction, we use the same CNN 
that is used for part detection. The  whole image is warped
 in order to use it as input for the CNN. The activations of a hidden layer are then used to build a feature vector.
For training as well as for testing, the part features are extracted from square shaped patches 
around the estimated part position.
The width and height of the patches are calculated as $p=\sqrt{n\cdot m \cdot \lambda}$, 
where $m$ and $n$ is the width and height of the image, respectively, and $\lambda$ 
is a parameter. We use $\lambda = 0.1$ in our experiments. 
Similar to the global feature extraction, the patches are resized in order
to use them as input for the CNN. The activations of a hidden layer for all patches
are then concatenated with the global feature. 
The resulting features are used to train a linear one-vs-all support vector machine for every category. 
We also tried various explicit kernel maps as demonstrated in \cite{goering2014parttransfer}, but 
none of these techniques lead to a performance gain.

\section{Experiments}
\label{sec:experiments}

\myparagraph{Experimental Setup}
We evaluate our approach on Caltech Birds CUB200-2011~\cite{WahCUB_200_2011},
a challenging dataset for fine-grained classification. 
The dataset consists of 11788 labeled photographs of 200 bird species in their natural environment. 
Further qualitative results are presented on the Columbia dogs dataset~\cite{liu12exemplarBasedDogParts}.
Besides class labels, both datasets come with semantic part annotations. The birds dataset also provides 
ground-truth bounding boxes. 

We use the CNN framework DeCAF~\cite{donahue2013decaf} and
the network learned on the ILSVRC 2012 dataset provided by the authors of \cite{donahue2013decaf}.
The CNN takes a $227\times227$ RGB color image as input and transforms it
into a vector of 1000 class scores. Details about the architecture of the network
are given in \cite{donahue2013decaf} as well as in \cite{krizhevsky2012imagenet} and we skip
the details here, because our approach can be used with any CNN.
For all experiments, we use the output of the last pooling layer to calculate
the gradient maps with respect to the input image. It consists of 256 
channels with 36 elements each and directly follows the last convolutional layer.
For each gradient map, we use the GMM method as explained in \sectionname~\ref{sec:part_detection}
with $K=2$ components, a maximum of 100 EM iterations, and three repetitions
with random initialization to increase robustness.

In the classification experiments, the same CNN model as for 
the part detection is used. The activations of the last hidden
layer are taken as a feature vector. For the part-based classification, 
the learned part detectors are used. From the estimated part positions 
of the training and test images, squared patches  
are extracted using $\lambda=0.1$. Each patch is then warped to size $227\times 227$ in order to be used
as input for the CNN and the activations of the last hidden layer are used as a feature
vector for this part. 

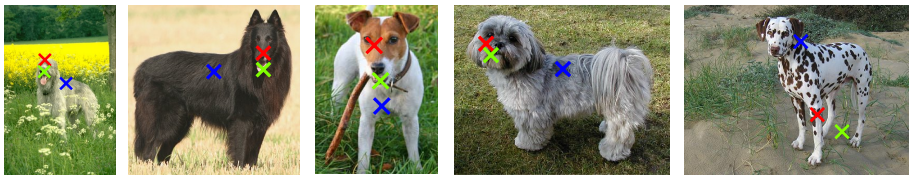
\begin{figure}[tbp]
\begin{center}
 
\resizebox{\textwidth}{!}{%
\resizebox{!}{2cm}{%
\begin{centering}
%
%
%
\definecolor{mycolor1}{rgb}{0.50000,1.00000,0.00000}%
\definecolor{mycolor2}{rgb}{0.00000,0.00000,1.0}%
\begin{tikzpicture}

\begin{axis}[%
width=3.63333333333333in,
height=5.55555555555556in,
axis on top,
scale only axis,
xmin=0.5,
xmax=327.5,
y dir=reverse,
ymin=0.5,
ymax=500.5,
hide axis
]
\addplot [forget plot] graphics [xmin=0.5,xmax=327.5,ymin=0.5,ymax=500.5] {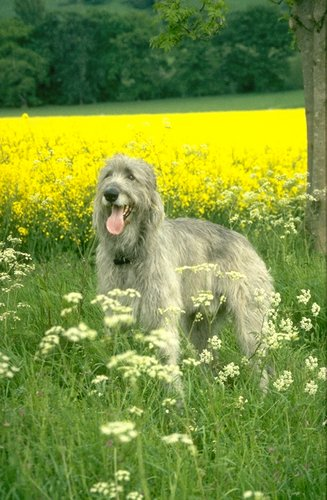};
\addplot [color=red,line width=6.0pt,mark size=20.0pt,only marks,mark=x,mark options={solid},forget plot]
  table[row sep=crcr]{119	161\\
};
\addplot [color=mycolor1,line width=6.0pt,mark size=20.0pt,only marks,mark=x,mark options={solid},forget plot]
  table[row sep=crcr]{117	196\\
};
\addplot [color=mycolor2,line width=6.0pt,mark size=20.0pt,only marks,mark=x,mark options={solid},forget plot]
  table[row sep=crcr]{182	231\\
};
\end{axis}
\end{tikzpicture}%
\end{centering}
}
\resizebox{!}{2cm}{%
\begin{centering}
%
%
%
\definecolor{mycolor1}{rgb}{0.50000,1.00000,0.00000}%
\definecolor{mycolor2}{rgb}{0.00000,0.00000,1.0}%
\begin{tikzpicture}

\begin{axis}[%
width=3.66666666666667in,
height=3.62222222222222in,
axis on top,
scale only axis,
xmin=0.5,
xmax=330.5,
y dir=reverse,
ymin=0.5,
ymax=326.5,
hide axis
]
\addplot [forget plot] graphics [xmin=0.5,xmax=330.5,ymin=0.5,ymax=326.5] {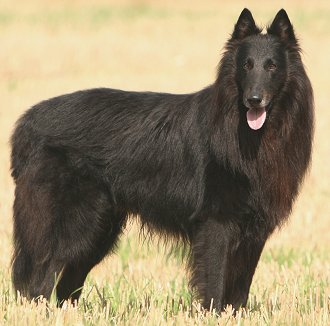};
\addplot [color=red,line width=5.0pt,mark size=15.0pt,only marks,mark=x,mark options={solid},forget plot]
  table[row sep=crcr]{261	92\\
};
\addplot [color=mycolor1,line width=5.0pt,mark size=15.0pt,only marks,mark=x,mark options={solid},forget plot]
  table[row sep=crcr]{261	124\\
};
\addplot [color=mycolor2,line width=5.0pt,mark size=15.0pt,only marks,mark=x,mark options={solid},forget plot]
  table[row sep=crcr]{165	128\\
};
\end{axis}
\end{tikzpicture}%
\end{centering}
}
\resizebox{!}{2cm}{%
\begin{centering}
%
%
%
\definecolor{mycolor1}{rgb}{0.50000,1.00000,0.00000}%
\definecolor{mycolor2}{rgb}{0.00000,0.00000,1.0}%
\begin{tikzpicture}

\begin{axis}[%
width=2.15555555555556in,
height=2.95555555555556in,
axis on top,
scale only axis,
xmin=0.5,
xmax=194.5,
y dir=reverse,
ymin=0.5,
ymax=266.5,
hide axis
]
\addplot [forget plot] graphics [xmin=0.5,xmax=194.5,ymin=0.5,ymax=266.5] {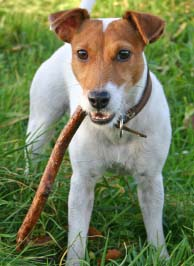};
\addplot [color=red,line width=4.0pt,mark size=14.0pt,only marks,mark=x,mark options={solid},forget plot]
  table[row sep=crcr]{94	63\\
};
\addplot [color=mycolor1,line width=4.0pt,mark size=14.0pt,only marks,mark=x,mark options={solid},forget plot]
  table[row sep=crcr]{104	119\\
};
\addplot [color=mycolor2,line width=4.0pt,mark size=14.0pt,only marks,mark=x,mark options={solid},forget plot]
  table[row sep=crcr]{106	159\\
};
\end{axis}
\end{tikzpicture}%
\end{centering}
}
\resizebox{!}{2cm}{%
\begin{centering}
%
%
%
\definecolor{mycolor1}{rgb}{0.50000,1.00000,0.00000}%
\definecolor{mycolor2}{rgb}{0.00000,0.00000,1.0}%
\begin{tikzpicture}

\begin{axis}[%
width=4.44444444444444in,
height=3.47777777777778in,
axis on top,
scale only axis,
xmin=0.5,
xmax=400.5,
y dir=reverse,
ymin=0.5,
ymax=313.5,
hide axis
]
\addplot [forget plot] graphics [xmin=0.5,xmax=400.5,ymin=0.5,ymax=313.5] {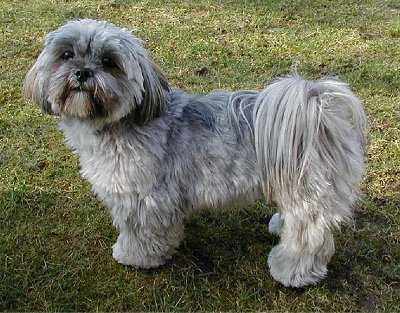};
\addplot [color=red,line width=5.0pt,mark size=15.0pt,only marks,mark=x,mark options={solid},forget plot]
  table[row sep=crcr]{60	71\\
};
\addplot [color=mycolor1,line width=5.0pt,mark size=15.0pt,only marks,mark=x,mark options={solid},forget plot]
  table[row sep=crcr]{69	92\\
};
\addplot [color=mycolor2,line width=5.0pt,mark size=15.0pt,only marks,mark=x,mark options={solid},forget plot]
  table[row sep=crcr]{203	118\\
};
\end{axis}
\end{tikzpicture}%
\end{centering}
}
\resizebox{!}{2cm}{%
\begin{centering}
%
%
%
\definecolor{mycolor1}{rgb}{0.50000,1.00000,0.00000}%
\definecolor{mycolor2}{rgb}{0.00000,0.00000,1.0}%
\begin{tikzpicture}

\begin{axis}[%
width=5.55555555555556in,
height=4.16666666666667in,
axis on top,
scale only axis,
xmin=0.5,
xmax=500.5,
y dir=reverse,
ymin=0.5,
ymax=375.5,
hide axis
]
\addplot [forget plot] graphics [xmin=0.5,xmax=500.5,ymin=0.5,ymax=375.5] {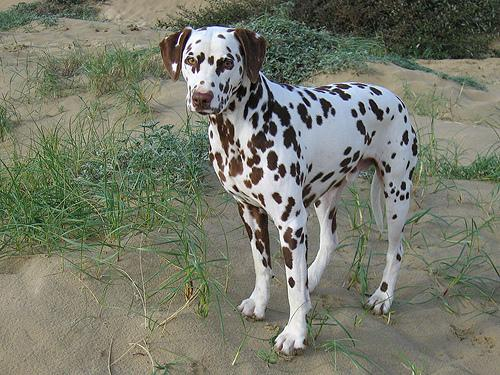};
\addplot [color=red,line width=5.0pt,mark size=17.0pt,only marks,mark=x,mark options={solid},forget plot]
  table[row sep=crcr]{297	243\\
};
\addplot [color=mycolor1,line width=5.0pt,mark size=17.0pt,only marks,mark=x,mark options={solid},forget plot]
  table[row sep=crcr]{352	281\\
};
\addplot [color=mycolor2,line width=5.0pt,mark size=17.0pt,only marks,mark=x,mark options={solid},forget plot]
  table[row sep=crcr]{260	80\\
};
\end{axis}
\end{tikzpicture}%
\end{centering}
}
}
\caption{Detections of the discovered part detectors on the dogs dataset. Green corresponds 
to the nose, red to the head and blue to the body of the dog. The last image 
is a failure case. Best viewed in color.\label{fig:part_qualitative_results_dog} }

\end{center}

\end{figure}

\begin{figure}[tbp]
\CenterFloatBoxes
\begin{floatrow}
  \ttabbox{%
    {\small
      \begin{tabular}{lc}
      \toprule 
      Method  & Norm. Error\tabularnewline
      \midrule
      Ours (GMM, BB)  & \textbf{0.16} \tabularnewline
	Ours (GMM, Full)   & 0.17 \tabularnewline
	Ours (MaxG, BB) & 0.17 \tabularnewline
	 Part Transfer \cite{goering2014parttransfer} (BB)& 0.18\tabularnewline
	 Ours (CNN from scratch)& 0.36\tabularnewline
      \bottomrule 
      \end{tabular}
    }
  }{%
    \caption{Part localization error on the CUB-2011-200 dataset for our \textit{part strategy} method w/ and 
    w/o GMM for finding the activation centers, our method w/ and w/o restricting the localization
    to the bounding box (BB), and the method of~\cite{goering2014parttransfer} \label{tab:Part_Channel_Assoc}. 
    In addition, we also show the performance of our approach for a CNN trained from scratch on CUB200-2011. }
  }
  \ffigbox{%
  \resizebox{0.5\textwidth}{!}{%
  \resizebox{!}{1cm}{%
  \begin{centering}
%
%
%
\definecolor{mycolor1}{rgb}{0.00000,0.75000,0.75000}%
\definecolor{mycolor2}{rgb}{0.75000,0.00000,0.75000}%
\begin{tikzpicture}

\begin{axis}[%
width=5.55555555555556in,
height=4.06666666666667in,
axis on top,
scale only axis,
xmin=0.5,
xmax=500.5,
y dir=reverse,
ymin=0.5,
ymax=366.5,
hide axis
]
\addplot [forget plot] graphics [xmin=0.5,xmax=500.5,ymin=0.5,ymax=366.5] {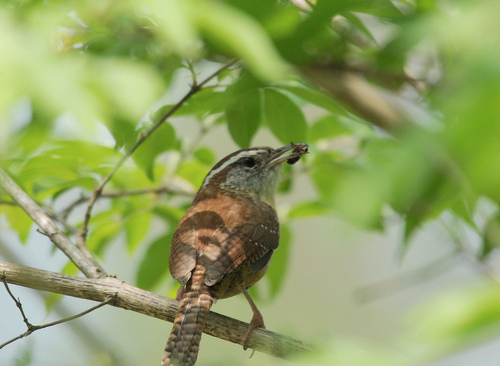};
\addplot [color=blue,line width=7.5pt,mark size=20.0pt,only marks,mark=x,mark options={solid},forget plot]
  table[row sep=crcr]{243	187\\
};
\addplot [color=red,line width=7.5pt,mark size=20.0pt,only marks,mark=x,mark options={solid},forget plot]
  table[row sep=crcr]{281	158\\
};
\addplot [color=white!99!black,line width=7.5pt,mark size=20.0pt,only marks,mark=x,mark options={solid},forget plot]
  table[row sep=crcr]{257	318\\
};
\addplot [color=mycolor1,line width=7.5pt,mark size=20.0pt,only marks,mark=x,mark options={solid},forget plot]
  table[row sep=crcr]{220	164\\
};
\addplot [color=mycolor2,line width=7.5pt,mark size=20.0pt,only marks,mark=x,mark options={solid},forget plot]
  table[row sep=crcr]{204	303\\
};
\end{axis}
\end{tikzpicture}%
  \end{centering}
  }
  \resizebox{!}{1cm}{%
  \begin{centering}
%
%
%
\definecolor{mycolor1}{rgb}{0.00000,0.75000,0.75000}%
\definecolor{mycolor2}{rgb}{0.75000,0.00000,0.75000}%
\begin{tikzpicture}

\begin{axis}[%
width=5.55555555555556in,
height=2.92222222222222in,
axis on top,
scale only axis,
xmin=0.5,
xmax=500.5,
y dir=reverse,
ymin=0.5,
ymax=263.5,
hide axis
]
\addplot [forget plot] graphics [xmin=0.5,xmax=500.5,ymin=0.5,ymax=263.5] {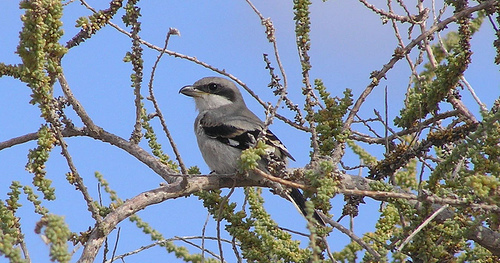};
\addplot [color=blue,line width=5.6pt,mark size=15.0pt,only marks,mark=x,mark options={solid},forget plot]
  table[row sep=crcr]{232	126\\
};
\addplot [color=red,line width=5.6pt,mark size=15.0pt,only marks,mark=x,mark options={solid},forget plot]
  table[row sep=crcr]{202	94\\
};
\addplot [color=white!99!black,line width=5.6pt,mark size=15.0pt,only marks,mark=x,mark options={solid},forget plot]
  table[row sep=crcr]{222	175\\
};
\addplot [color=mycolor1,line width=5.6pt,mark size=15.0pt,only marks,mark=x,mark options={solid},forget plot]
  table[row sep=crcr]{238	132\\
};
\addplot [color=mycolor2,line width=5.6pt,mark size=15.0pt,only marks,mark=x,mark options={solid},forget plot]
  table[row sep=crcr]{233	163\\
};
\end{axis}
\end{tikzpicture}%
  \end{centering}
  }
  }
  \vspace{-0.2cm}\\
  \resizebox{0.5\textwidth}{!}{%
  \resizebox{!}{1cm}{%
  \begin{centering}
%
%
%
\definecolor{mycolor1}{rgb}{0.00000,0.75000,0.75000}%
\definecolor{mycolor2}{rgb}{0.75000,0.00000,0.75000}%
\begin{tikzpicture}

\begin{axis}[%
width=5.55555555555556in,
height=4.16666666666667in,
axis on top,
scale only axis,
xmin=0.5,
xmax=500.5,
y dir=reverse,
ymin=0.5,
ymax=375.5,
hide axis
]
\addplot [forget plot] graphics [xmin=0.5,xmax=500.5,ymin=0.5,ymax=375.5] {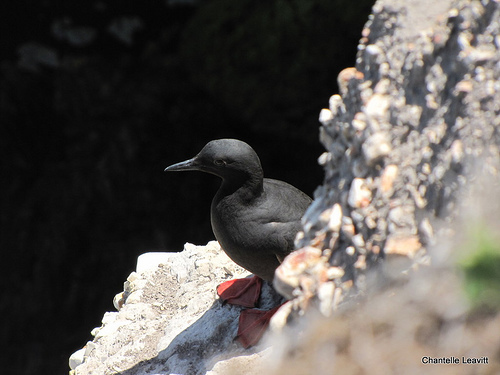};
\addplot [color=blue,line width=6.0pt,mark size=18.0pt,only marks,mark=x,mark options={solid},forget plot]
  table[row sep=crcr]{214	175\\
};
\addplot [color=red,line width=6.0pt,mark size=18.0pt,only marks,mark=x,mark options={solid},forget plot]
  table[row sep=crcr]{228	162\\
};
\addplot [color=white!99!black,line width=6.0pt,mark size=18.0pt,only marks,mark=x,mark options={solid},forget plot]
  table[row sep=crcr]{257	284\\
};
\addplot [color=mycolor1,line width=6.0pt,mark size=18.0pt,only marks,mark=x,mark options={solid},forget plot]
  table[row sep=crcr]{236	197\\
};
\addplot [color=mycolor2,line width=6.0pt,mark size=18.0pt,only marks,mark=x,mark options={solid},forget plot]
  table[row sep=crcr]{269	225\\
};
\end{axis}
\end{tikzpicture}%
  \end{centering}
  }
  \resizebox{!}{1cm}{%
  \begin{centering}
%
%
%
\definecolor{mycolor1}{rgb}{0.00000,0.75000,0.75000}%
\definecolor{mycolor2}{rgb}{0.75000,0.00000,0.75000}%
\begin{tikzpicture}

\begin{axis}[%
width=5.55555555555556in,
height=4.16666666666667in,
axis on top,
scale only axis,
xmin=0.5,
xmax=500.5,
y dir=reverse,
ymin=0.5,
ymax=375.5,
hide axis
]
\addplot [forget plot] graphics [xmin=0.5,xmax=500.5,ymin=0.5,ymax=375.5] {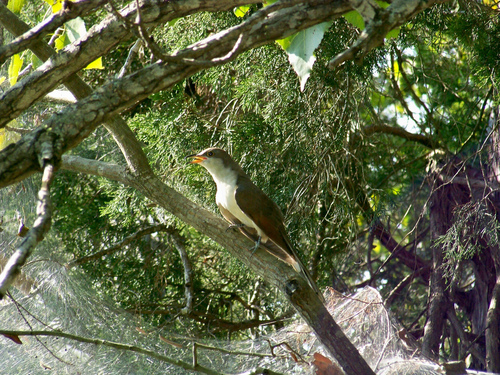};
\addplot [color=blue,line width=6.5pt,mark size=18.0pt,only marks,mark=x,mark options={solid},forget plot]
  table[row sep=crcr]{371	329\\
};
\addplot [color=red,line width=6.5pt,mark size=18.0pt,only marks,mark=x,mark options={solid},forget plot]
  table[row sep=crcr]{205	157\\
};
\addplot [color=white!99!black,line width=6.5pt,mark size=18.0pt,only marks,mark=x,mark options={solid},forget plot]
  table[row sep=crcr]{262	251\\
};
\addplot [color=mycolor1,line width=6.5pt,mark size=18.0pt,only marks,mark=x,mark options={solid},forget plot]
  table[row sep=crcr]{107	97\\
};
\addplot [color=mycolor2,line width=6.5pt,mark size=18.0pt,only marks,mark=x,mark options={solid},forget plot]
  table[row sep=crcr]{266	225\\
};
\end{axis}
\end{tikzpicture}%
  \end{centering}
  }
  }
  }{%
\caption{Part localization examples from birds dataset. 
No bounding box and no geometric constraints for the part locations are used during the 
localization. The first three images are analyzed correctly, 
the belly and wing position are wrongly located in the fourth image. Best viewed in color.\label{fig:part_qualitative_results} }
  }
\end{floatrow}
\end{figure}

\myparagraph{Qualitative Evaluation}
Figures \ref{fig:part_qualitative_results_dog} and \ref{fig:part_qualitative_results} present some examples 
of our part localization applied to uncropped
test images. For both datasets, relevant channels were identified using the ground-truth part annotations.
In case of the dogs dataset, the three discovered detectors seem to correspond to the nose (red), head (green), and body (blue). 
Especially the nose is identified reliably. For the birds dataset, we present four challenging examples in 
which only a small portion of the image
contains the actual bird. In addition to this, the bird  is often partially occluded. Nevertheless,
our approach is able to precisely locate the head (red) and the legs (white). While there 
is more variance in the body (blue), wing (light blue), and tail (purple) location, they are still 
close to the real locations. The last example in \figurename~\ref{fig:part_qualitative_results_dog} and \figurename~\ref{fig:part_qualitative_results} shows a failure case.

  \begin{table}[tbp]
  \centering
  \begin{tabular}{ccclc}
  \toprule
  Training     &\multicolumn{2}{c}{Testing}  & Method & Recognition Rate\tabularnewline
  Parts        & BBox        & Parts   &        & \tabularnewline
  \midrule
  GT&GT& &Baseline (BBox CNN features)& 56.00\%\tabularnewline
  GT&GT&est.&Berg \etal~\cite{berg2013partclass}& 56.78\%\tabularnewline
  GT&GT&est.&Goering \etal~\cite{goering2014parttransfer}& 57.84\%\tabularnewline
  GT&GT&est.&Chai \etal~\cite{chai2013symbiotic}& 59.40\%\tabularnewline
  GT&GT&est.&Ours (\textit{part strategy}) & 62.53\%\tabularnewline
  GT&GT&GT&Ours (\textit{part strategy}) & 62.67\%\tabularnewline
  GT&GT&est.&Donahue \etal~\cite{donahue2013decaf}& 64.96\%\tabularnewline
  \midrule
  GT& &est.&Ours (\textit{part strategy}) & 60.17\%\tabularnewline
  GT& &GT&Ours (\textit{part strategy}) & 60.55\%\tabularnewline
  \midrule
   & & &Baseline (global CNN features) & 41.60\%\tabularnewline
   & &est.&Ours (\textit{counting}) & 51.93\%\tabularnewline
   & &est.&Ours (\textit{bounding box strategy}) & 53.75\%\tabularnewline
  \bottomrule
  \end{tabular}
    \caption{Species categorization performance on the CUB200-2011 dataset.
    We distinguish between different experimental setups, depending on whether the ground-truth parts
    are used in training and whether the ground-truth bounding box is used during testing\protect\footnotemark[1]. 
    \label{tab:finegrained}}
  \end{table}
\myparagraph{Evaluating the Part Localization Error}
First, we are interested to what extent the learned part detectors relate to 
 semantic parts.
After identifying the spatially most related channel for each semantic part, we can
apply our method to the test images to predict the location of semantic parts.
The localization errors are given in Table~\ref{tab:Part_Channel_Assoc}. 
For calculating the localization error, we follow the work of \cite{goering2014parttransfer} and 
use the mean pixel error normalized by the length of the diagonal of the ground-truth bounding box.

Our method achieves a significantly lower part localization error compared to \cite{goering2014parttransfer}
and a baseline using a CNN learned from scratch on the CUB200-2011 dataset only.
A detailed analysis of the part localization error for each part is given in the supplementary materials
including several observations about the channel-part correspondences: there are groups of parts that are
associated with the same channel and there are parts, such as the beak and the throat,
where we are twice as accurate as \cite{goering2014parttransfer}. 
This indicates that the system can distinguish different bird body parts without direct training, which is a surprising fact given that we use  a
CNN trained for a completely different task. 

\footnotetext[1]{After submission, two additional publications \cite{zhang14partRCNN,branson14cub75acc} 
    were published reporting 73.9\% and 75.7\% accuracy if part annotations are used only in training and 
    no ground-truth bounding box is used during testing. Both of these works perform fine-tuning of the CNN models, which is not done in our case.}
\myparagraph{Evaluation for Part-based Fine-grained Classification}
We apply the presented part-based classification method to the CUB200-2011 dataset
in order to evaluate to what extend the predicted part locations contribute to 
a higher accuracy. 
As can be seen in Table~\ref{tab:finegrained}, our approach achieves a classification accuracy of 62.5\%, which is one
of the best results on CUB-2011-200 without fine-tuned CNNs.
Whereas the method of \cite{goering2014parttransfer} heavily relies on the ground-truth bounding box
for part estimation, our method can also perform fine-grained classification on unconstrained full images without a manual
preselection of the area containing the bird. These results are also shown in Table~\ref{tab:finegrained}.

    As the main focus of this paper is the part localization, two more results are
presented. First, the performance without using any part-based representation is
a lower bound of our approach, since we also add global features to our feature representation. 
The recognition rate in this case is 56.0\% for the constrained and 41.6\% for the unconstrained
setting. 
Second, the performance with ground-truth part locations is 
an upper bound, if we assume that human annotated semantic parts are also the best ones for automatic 
classification. In this case, the accuracy is 62.7\% and 60.6\%, respectively. 
The recognition rate of our approach with estimated part locations and the upper bound are nearly 
identical with a difference of only 0.2\% and 0.5\%, respectively.
We also show the results in the case no ground-truth part annotations are provided for the training data. 
A baseline is provided by using CNN activations of the uncropped image for classification. 
The presented \textit{counting strategy} as well as the \textit{bounding box strategy} are able to significantly outperform 
this baseline by over 10\% accuracy. The \textit{bounding box strategy} performs best with only 6\% less performance 
compared to the \textit{part strategy}, which makes use of ground-truth part locations during training.

\section{Conclusions}

We have presented a novel approach for object part discovery and 
detection with pre-trained deep models. The motivation
for our work was that deep convolutional neural network models 
learned on large-scale object databases such as ImageNet are already able to robustly
detect basic shapes typically found in natural images. We exploit this ability 
to discover useful parts for a fine-grained recognition task by analyzing gradient  
maps of deep models and selecting activation centers related to annotated semantic parts or bounding boxes.
After this simple learning step, part detection basically comes for free when 
applying the deep CNN to the image and detecting parts takes only a few seconds. 
Our experimental results show that in combination with a 
part-based classification approach this leads to 
an excellent performance of $62.5\%$ on the CUB-2011-200 dataset. 
In contrast to previous work~\cite{goering2014parttransfer}, 
our approach is also suitable for situations when the ground-truth bounding box is not given during testing. 
In this scenario, we obtain an accuracy of $60.1\%$, which is only slightly 
less than the result for the restricted setting with given bounding boxes. Furthermore, we also show
how to learn without given ground-truth part locations, making fine-grained recognition feasible without
huge annotation costs.

Future work will focus on dense features for the 
input image that can be obtained by stacking all gradient
maps of a layer. This allows us to apply previous part-based models like DPM, which also include geometric constraints
for the part locations. 
Furthermore, multiple channels can relate to 
the same part and our approach can be very easily modified to tackle these situations using 
an iterative approach. 

\myparagraph{Acknowledgments} 
This work was supported by Nvidia with a hardware donation.

\bibliographystyle{splncs}
\bibliography{paper}

\pagebreak

\title{Supplemental Materials: Part Detector Discovery in Deep Convolutional Neural Networks }
     
\titlerunning{Part Detector Discovery in Deep Convolutional Neural Networks}
\authorrunning{Marcel Simon \and Erik Rodner \and Joachim Denzler}

\author{Marcel Simon \and Erik Rodner \and Joachim Denzler}
\institute{Computer Vision Group, Friedrich Schiller University of Jena, Germany\\
\url{www.inf-cv.uni-jena.de}}

\maketitle

\setcounter{equation}{0}
\setcounter{figure}{0}
\setcounter{table}{0}
\setcounter{section}{0}
\makeatletter
\renewcommand{\theequation}{S\arabic{equation}}
\renewcommand{\thefigure}{S\arabic{figure}}
\renewcommand{\thetable}{S\arabic{table}}
\renewcommand{\thesection}{S\arabic{section}}

\section{Evaluating the Part Localization Error in Detail}
\definecolor{Gray}{gray}{0.93}%
\newcommand\helperrule{\arrayrulecolor{Gray}\midrule\arrayrulecolor{black}}%
\begin{table}[p]
\centering
\caption{Part localization error on the CUB-2011-200 dataset for our method w/ and 
w/o GMM for finding the activation centers, our method w/ and w/o restricting the localization
to the bounding box (BB), and the method of \cite{goering2014parttransfer}.
\label{tab:Part_Channel_Assoc_Detail}}
{\small
\begin{tabular}{lccccc}
\toprule 
Part & Channel & \multicolumn{4}{c}{Normalized Euclidean error}\tabularnewline
 &  & Ours & Ours & Ours & Part transfer\tabularnewline
 &  & (GMM, BB) & (GMM, Full) & (MaxG, BB) & \cite{goering2014parttransfer} (BB)\tabularnewline
\midrule
Back & 6 & 0.17 & 0.20 & 0.19 & \textbf{0.11}\tabularnewline
\helperrule 
Beak & 208 & 0.17 & 0.15 & \textbf{0.14} & 0.26\tabularnewline
\helperrule 
Belly & 171 & 0.14 & 0.17 & 0.17 & \textbf{0.13}\tabularnewline
\helperrule 
Breast & 208 & \textbf{0.15} & 0.18 & 0.21 & 0.20 \tabularnewline
\helperrule 
Crown & 208 & 0.16 & 0.16 & \textbf{0.13} & 0.20 \tabularnewline
\helperrule 
Forehead & 208 & 0.16 & 0.14 & \textbf{0.12} & 0.24\tabularnewline
\helperrule 
Left eye & 208 & 0.13 & 0.12 & \textbf{0.11} & 0.17\tabularnewline
\helperrule 
Left leg & 171 & 0.15 & 0.16 & 0.15 & \textbf{0.13}\tabularnewline
\helperrule 
Left wing & 73 & 0.18 & 0.20 & 0.22 & \textbf{0.15}\tabularnewline
\helperrule 
Nape & 208 & 0.15 & 0.16 & 0.17 & \textbf{0.13}\tabularnewline
\helperrule 
Right eye & 208 & 0.13 & 0.12 & \textbf{0.11} & 0.31 \tabularnewline
\helperrule 
Right leg & 171 & 0.16 & 0.16 & \textbf{0.15} & \textbf{0.15}\tabularnewline
\helperrule 
Right wing & 73 & 0.17 & 0.20 & 0.21 & \textbf{0.13}\tabularnewline
\helperrule 
Tail & 54 & 0.33 & 0.36 & 0.35 & \textbf{0.28}\tabularnewline
\helperrule
Throat & 208 & \textbf{0.10} & 0.11 & 0.13 & 0.23\tabularnewline
\midrule
Average & - & \textbf{0.16} & 0.17 & 0.17 & 0.18\tabularnewline
\bottomrule 
\end{tabular}
}
\end{table}

The best matching channel as well as the localization errors are given in Table~\ref{tab:Part_Channel_Assoc_Detail}. 
For calculating the localization error, we follow the work of \cite{goering2014parttransfer} and 
use the mean pixel error normalized by the length of the diagonal of the ground-truth bounding box.
Interestingly, there are groups of parts that are
associated with the same channel. The first group includes all parts
near the head: beak, crown, forehead, left and right eye,
breast, nape and throat. The second group is comprised of the left and right wing
and the third one is the belly
and the left and right leg. The back and the tail are each associated
with a separate channel. This indicates that the system can distinguish
different larger bird body parts, which is a surprising fact given that we use  a
CNN trained for a completely different task. 

\begin{figure}[p]
\centering
\resizebox{!}{0.45\linewidth}{
    \input{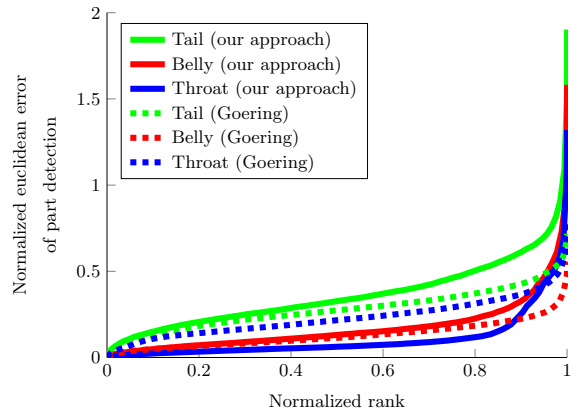}
}
\caption{Distribution of the part localization error: the sorted normalized localization errors for the tail, the belly,
and the throat compared to the approach of \cite{goering2014parttransfer} is shown.}
\label{fig:Part_Loc_Acc}
\end{figure}

In comparison to \cite{goering2014parttransfer}, we achieve a lower overall error 
with the benefit of our method most prominent in case of the beak and the throat.
In case of the throat, the presented approach is more than twice as accurate as the one of
\cite{goering2014parttransfer}. The semantic part with the highest localization
error is the tail, which seems to be challenging in general judging from the results of our competitor.
A detailed part localization error analysis for a few selected parts is given in \figurename~\ref{fig:Part_Loc_Acc}.
In this plot, the sorted normalized errors for all test images are
shown. The $x$-axis corresponds to the rank in the sorted error list,
and the $y$-axis is the normalized distance to the ground-truth part
location. The dotted lines are the results of \cite{goering2014parttransfer}
while the solid lines correspond to our approach.
In case of the throat, the error of the presented approach is significantly
lower and less than half in most cases. The other parts are located
with a comparable accuracy. 

\end{document}